\documentclass{article}

\usepackage[english]{babel}

\usepackage[letterpaper,top=2cm,bottom=2cm,left=3cm,right=3cm,marginparwidth=1.75cm]{geometry}

\usepackage{amsmath}
\usepackage{graphicx}
\usepackage[colorlinks=true, allcolors=blue]{hyperref}
\usepackage{bm}
\usepackage{comment}
\usepackage{enumitem}
\usepackage{subcaption}

\usepackage{csquotes}
\usepackage[backend=biber, natbib=true, style=alphabetic, maxbibnames=99]{biblatex}
\usepackage{xspace}

\newcommand*{\eg}{e.g.\@\xspace}
\newcommand*{\ie}{i.e.\@\xspace}

\newcommand*{\cf}{cf.\@\xspace}

\usepackage{amsthm}
\newtheorem{definition}{Definition}
\newtheorem{prop}{Proposition}
\newtheorem{example}{Example}
\newtheorem{remark}{Remark}
\newtheorem{lemma}{Lemma}

\newtheorem{corollary}{Corollary}

\usepackage{amssymb}
\usepackage{mathrsfs}

\DeclareMathOperator*{\argmax}{arg\,max}
\DeclareMathOperator*{\argmin}{arg\,min}

\newcommand{\TwoP}{2^{\Delta(\Z)}}

\newcommand{\X}{\mathcal X}
\newcommand{\Y}{\mathcal Y}
\newcommand{\Z}{\mathcal Z}
\renewcommand{\P}{\mathcal P}
\newcommand{\Q}{\mathcal Q}
\renewcommand{\L}{\mathcal L}
\newcommand{\LIP}{\mathscr L}
\newcommand{\scP}{\mathscr P}
\newcommand{\R}{\mathcal R}

\newcommand{\co}{\overline{\operatorname{co}}}
\newcommand{\ellce}{\ell^{\mathrm{CE}}}

\title{Property Elicitation on Imprecise Probabilities\\[1ex]%
\large Working Paper\footnote{This work was presented as a poster at the 14th International Symposium on Imprecise Probabilities: Theories and Application (\href{https://isipta25.sipta.org/home}{ISIPTA 2025}) in Bielefeld, Germany.}}

\author{James Bailie{\normalsize $^1$} and Rabanus Derr{\normalsize $^2$}\footnote{The authors are listed in alphabetical order of their last name.}}

\date{\normalsize $^1$Harvard University, Cambridge MA, USA \\
$^2$University of Tübingen and Tübingen AI Center, Tübingen, Germany \\[2ex]
\today}
\begin{document}
\maketitle

\begin{abstract}
    Property elicitation studies which attributes of a probability distribution can be determined by minimizing a risk. We investigate a generalization of property elicitation to imprecise probabilities (IP). This investigation is motivated by distributionally robust optimization and multi-distribution learning. Both those frameworks replace the minimization of a single risk over a (precise) probability by a maximin risk minimization over a set of probabilities -- i.e. an IP. We show what \emph{can} be learned in those multi-distribution setups by providing necessary and sufficient conditions for the elicitability of an IP-property. Central to these conditions is the observation made in related literature that the elicited IP-property is the corresponding classical property of the probability in the IP with the maximum Bayes risk.
    
\end{abstract}

\section{Motivation: What \emph{can} be ``minimax-learned'' in a multi-distribution setup?}

Machine learning in many of its forms is, at its core, minimizing an expected loss (\ie a risk). 
Interestingly, expected loss minimization is also central to another strand of literature: \emph{property elicitation}. 
Starting with the seminal work of \citeauthor{osband1985providing} \cite{osband1985providing}, this field takes a distinct view of loss minimization: It mainly asks that given a property---i.e. a function of the data generating distribution---does there exist a loss function which elicits the property---i.e. the property is the minimizer of the expected loss under a certain distribution.

A classical example of an elicitable property is the mean. Minimizing the expected squared error elicits the mean of the data generating distribution. Furthermore, all quantiles of a distribution are elicitable through appropriate parametrizations of the pinball loss. On the other hand, the conditional value at risk (sometimes called expected shortfall) of a distribution is not elicitable \citep{gneiting2011making, ziegel2016coherence}.

To put the questions described above more succinctly: the field of property elictation aims to answer what \emph{can be} elicited through minimizing an expected loss. Those with a statistical background will immediately see that this question is equivalent to what properties are possible to learn by $M$-estimation. In recent years, these questions have largely been answered by the field, culminating in the characterization of elicitable properties (see \cite[Theorem 1]{lambert2019elicitation} and \cite{steinwart2014elicitation}).

However, over the last decades the risk minimization framework has grown into an entire family of related learning paradigms, many of which deviate from the classical setup considered in the property elicitation literature. As motivation for the present paper, we highlight two such paradigms: \emph{multi-distribution learning} (MDL) \citep{haghtalab2022demand} and \emph{distributionally robust optimization} (DRO) \citep{rahimianFrameworksResultsDistributionally2022, kuhn2025distributionally}.

Instead of minimizing an expected loss, MDL replaces the expectation operator by the supremum over a set of expectations, following the rationale that during training the actual distribution of interest---from which data will be drawn in deployment---is unknown. Rather, good prediction performance should be guaranteed across the class of distributions over which the supremum is taken. In other words, MDL can be described as expected loss minimization over an \emph{imprecise probability} (IP) \citep{augustin2014introduction}, rather than over a single (precise) probability. 

A closely related framework is \emph{distributionally robust optimization} (DRO) \citep{rahimianFrameworksResultsDistributionally2022, kuhn2025distributionally}, which is an IP generalization of classical stochastic programming. It addresses scenarios where the data generating distribution is not known precisely, but rather, may be adversarially chosen from a family of possible distributions---the ``ambiguity set''---which are consistent with the existing knowledge about the data. As such, DRO optimizes the worst-case risk taken over this imprecise probability.

In this work, we address a fundamental question motivated by the MDL and DRO paradigms:
\begin{center}
    \textbf{What \emph{can} be ``minimax-learned'' in a multi-distribution setup?}
\end{center}
Essentially, we reiterate the central question of the field of property elicitation---but we are now concerned with loss minimization over IP-expectations, rather than standard expectations. 

We introduce IP-properties and a definition of elicitability related to the $\Gamma$-minimax decision rule (\cf \citep{troffaes2007decision}). Then, we provide a series of necessary conditions for IP-elicitability. We give a first sufficient condition for elicitability of an IP-property. We conclude the work with open questions.

\section{Related work}
As noted in the introduction, property elicitation was initiated in \cite{osband1985providing}, although it has its roots in the classical statistical theory of $M$-estimation. Our work extends the concept of property elicitation to imprecise probabilities.

The term ``elicitation'' often refers to the problem of recovering the true data generating distribution, for example by minimizing the expected value of a proper scoring rule in the context of forecasting \citep{brierVerificationForecastsExpressed1950}. Property elicitation can be viewed as a generalization of this idea, as it considers whether a property of the data generating distribution 
can be ascertained. This property can be the distribution itself (this would recover the classical elicitation problem), or only an attribute of the distribution, such as its mean or variance. 

Recently, the study of proper scoring rules for imprecise probabilities has gained new interest \citep{konek2023evaluating, frohlich2024scoring, singh2025truthful, schervish2025elicitation}. Although this paper is not explicitly concerned with proper scoring rules, it is connected to this new work---most closely with \cite{frohlich2024scoring, singh2025truthful} and \citep{schervish2025elicitation}. Those works share that they encompass the elicitation of probability distributions via the $\Gamma$-maximin decision rule, which is central for our definition of IP elicitability (Remark~\ref{remarkGammaMinimax}). 

Our work also differs from \cite{gneiting2011making, fissler2021forecast, brehmer2021scoring} who consider the elicitation of ``imprecise'' interval valued (or set-valued) properties on precise probability distributions.

Properties of an IP can also be understood as structural statements about the IP, \eg, whether the IP is 2-monotone or whether it has a set of desirable gambles which is invariant to certain transformations \citep{mirandaStructuralJudgements2014}. Clearly, these binary statements are representable as properties following our definition. However, it is not clear whether these properties are elicitable in the sense we present in this paper.

Finally, DRO and MDL literature is most often concerned with the computation of minimax-optimal learners \citep{haghtalab2022demand, rahimianFrameworksResultsDistributionally2022, kuhn2025distributionally}. The tools used, \eg, minimax theorems, are similar to ours (\cf Proposition~\ref{prop:necessary condition of IP-elicitability via Bayes pair}). However, we focus on the question of what, in principle, can be learned using a minimax-objective as standard in DRO or MDL.

\section{Background: Elicitable properties of probabilities}
Let us first recapitulate some background on property elicitation. We follow the presentation of \citep{gneiting2011making}. Fix some measurable space $(\Z, \Sigma)$. Let $\Delta(\Z)$ denote the set of probabilities on $\Z$ (the ``probability simplex'' of $\Z$). Given a set $A$, we denote its power set by $2^A$.
\begin{definition}[Elicitable property]
\label{def:elicitable property}
    Let $\R \subseteq \mathbb R^p$ be some set of property values.
    A \emph{property} is a function $f : \mathbb P \to 2^\R$ for some $\mathbb P \subseteq \Delta(\Z)$. For $\theta \in \R$, the \emph{level set} $\L_\theta$ of a property is defined as
    \begin{align*}
        \L_\theta = \{P \in \mathbb P \colon \theta \in f(P) \},
    \end{align*}
    (Without loss of generality we assume throughout that $\R = \cup_{P \in \mathbb{P}} f(P)$.)
    
    A property $f$ is \emph{elicitable} if there exists a loss function $\ell : \R \times \Z \to \mathbb R$ such that for all $P \in \mathbb P$,
    \[f(P) = \argmin_{\theta \in \R} \mathbb E_{Z \sim P} [ \ell(\theta, Z) ].\]
\end{definition}
Note that we do \emph{not} neglect the conditional relationship between inputs $X$ and outputs $Y$ which is central to machine learning. However, we abstract away this distinction by using an encompassing random variable $Z$ which can be defined as (but is not restricted to) $Z = (X,Y)$ (\cf Example~\ref{ex:expected risk minimization in ml}).

It will typically be the case that $f(P)$ is a singleton set---\ie usually a property $f$ can be considered as a map $\mathbb P \to \R$. However, restricting ourselves to point-valued properties poses a complication: For $f : \mathbb P \to \R$ to be elicitable by the loss $\ell$, we would have to assume that the minimizer of the risk $\mathbb E_{Z \sim P} [ \ell(\theta, Z)$ is unique. To avoid making this assumption, we instead allow properties to be set-valued. Intuitively, this is akin to bundling together all the point-valued properties which are elicited by a non-uniquely-minimized loss. 

We also highlight that a property may be defined only on a subset $\mathbb P$ of the probability simplex $\Delta(\Z)$. This is useful because there are many properties (\eg the mean) that are not defined for all probability distributions.

The following example demonstrates the link between expected loss minimization---which is central to a large part of machine learning---and elicitability:
\begin{example}
\label{ex:expected risk minimization in ml}
    Let $\Z = \X \times \Y$ for some input set $\X \subseteq \mathbb{R}^d$ and some output set $\Y = \{ 1, \ldots, K\}$. Much of machine learning is concerned with finding a ``predictor''---i.e., a measurable function $g_\theta : \X \to \Delta (\Y)$ parametrized by $\theta \in \R$---which explains the relationship between inputs and outputs. A common way to find such a $g_\theta$ is via minimizing a loss---say, for example, the instance-wise cross-entropy loss $\ellce(\theta, z) = -\log(g_\theta^y(x))$, where $g_\theta^y(x)$ denotes the $y$-component of $g_\theta(x)$. Given a data generating distribution $P$, the parameter $\theta^*$ is optimal if 
    \begin{equation*}
    \label{eq:ML minimizer}
        \theta^* \in \argmin_{\theta \in \R} \mathbb{E}_{Z \sim P}[\ellce(\theta, Z)] = \argmin_{\theta \in \R} \mathbb{E}_{(X,Y) \sim P}[-\log(g_\theta^Y(X))]\,.
    \end{equation*}
    Observe that the property elicited by $\ell$ is the function $\Theta : \mathbb P \to 2^{\R}$ mapping a distribution $P$ to the set of optimal parameters under $P$. Here the domain $\mathbb P$ of $\Theta$ is the set of distributions for which there exists a minimizer of $\mathbb{E}_{Z \sim P}[\ellce(\theta, Z)]$.
\end{example}

We now introduce a special type of property pair, which will be of use in later discussion.
\begin{definition}[Bayes pair \cite{embrechts2021bayes}]
    Let $\ell \colon \R \times \Z \to \mathbb R$ be a loss function. The property pair $(\Theta, L_\Theta)$ is called a \emph{Bayes pair} if $\Theta$ is the property elicited by $\ell$ on $\mathbb{P}$ and, for all $P \in \mathbb P$,
    \begin{align*}
        L_\Theta(P) = \min_{\theta \in \R} \mathbb{E}_{Z \sim P}[\ell(\theta, Z)]\,.
    \end{align*}
\end{definition}
A Bayes pair $(\Theta, L_{\Theta})$ has special structure: $L_\Theta(P)$ is the minimum risk (what is called the Bayes risk in learning theory) and $\Theta(P)$ the set of risk minimizers. Note that $L_\Theta$ is elicitable on level sets of $\Theta$ \citep[Theorem 3]{frongillo2021elicitation}. While the minimizer is central to forecasting \citep{gneiting2011making}, the minimum risk is of interest in mathematical finance \citep{embrechts2021bayes}.
\begin{example}
\label{Bayes pair for squared loss}
    Let $\Z \subseteq \mathbb{R}$ and $\R \subseteq \Z$. Suppose that $\mathbb P$ is the set of all distributions with finite second moments. Let $\ell(\theta, z) = (\theta - z)^2$ be the loss function. For $P \in \mathbb P$, the property $\Theta(P)$ of the corresponding Bayes pair is the mean of $P$,
    because
    \begin{align*}
        \mathbb{E}_{Z \sim P}[(\theta-Z)^2] = \theta^2 -2 \theta \mathbb{E}_{Z \sim P}[Z] + \mathbb{E}_{Z \sim P}[Z^2]\,,
    \end{align*}
    is minimized by $\theta = \mathbb{E}_{Z \sim P}[Z]$. The corresponding $L_\Theta(P)$ is the variance of $P$ since
    \begin{align*}
        \min_{\theta \in \R} \mathbb{E}_{Z \sim P}[\ell(\theta, Z)^2] = \mathbb{E}_{Z \sim P}[(\mathbb{E}_{Z \sim P}[Z]-Z)^2]\,.
    \end{align*}
\end{example}

\section{Elicitable IP-properties via minimax}
As alluded to in the introduction, an IP is a set of probability distributions---\ie, a subset of $\Delta(\Z)$. IPs have found use in a wide variety of scenarios where a single (\ie, ``precise'') probability distribution is inadequate for modeling uncertainty \citep{augustin2014introduction}. Given the introductory remarks in the previous section, we propose the following definition for an IP-property. 
\begin{definition}[Elicitable IP-property]
\label{def:elicitable IP-property}
    Let $\R \subseteq \mathbb R^p$ be a set of property values.
    An \emph{IP-property} is a function $f : \scP \to 2^\R$ for some $\scP \subseteq \TwoP$ with $\emptyset \notin \scP$. For $\theta \in \R$, the \emph{level set} $\LIP_{\theta}$ of a property is defined as
    \begin{align*}
        \LIP_{\theta} = \{\P \in \scP :  \theta \in f(\P) \}\,,
    \end{align*}
    (As for precise probabilities, we assume without loss of generality that $\R = \cup_{\P \in \scP} f(\P)$.)
    
    An IP-property $f$ is \emph{elicitable} if there exists a loss function $\ell : \R \times \Z \to \mathbb R$ such that, for all $\P \in \scP$, 
    \begin{align*}
        f(\P) = \argmin_{\theta \in \R} \sup_{P \in \P} \mathbb{E}_{Z \sim P}[\ell(\theta, Z)]\,.
    \end{align*}
\end{definition}

As with precise properties, we allow an IP-property to be defined only on a subset $\scP$ of the set $\TwoP$ of all IPs. Nevertheless, as we show in the following lemma, elicitable IP-properties can be defined on $\TwoP$ if we make assumptions that ensure a minimizer of the relevant loss function exists. Under additional assumptions, we can also guarantee that this minimizer is unique. (These conditions are satisfied when, for instance, $\Z \subseteq \mathbb R^p$; $\R$ is a closed box in $\mathbb R^p$ and $\ell(\theta, z) = \|\theta-z\|_2^2$.)
\begin{lemma}[Existence and uniqueness of elicitation values]
\label{lemma:Existence and Uniqueness of Elicitation Value}
    If $\R$ is compact, and $\ell$ lower semi-continuous in $\theta$, then a (potentially non-unique) minimum of $\theta \mapsto \sup_{P \in \mathcal P} \mathbb E_{Z \sim P} [ \ell (\theta, Z) ]$ is attained for some $\theta^* \in \R$.
    If furthermore $\R$ is convex and $\ell \colon \R \times \Z \to \mathbb R$ strictly convex in $\R$ for all $z \in \Z$, then the minimizer $\theta^* \in \R$ is unique.
\end{lemma}
\begin{proof}
    The results follows from the fact that a point-wise supremum over a lower semi-continuous function is lower-semi-continuous \cite[15.23.d]{schechter1997handbook}, and that a minimum is then attained if this function ranges over a compact set  \cite[17.7.i]{schechter1997handbook}.
    We show the second statement by contradiction. Suppose that $\R$ is convex, $\ell$ strictly convex and $\theta^* \in \R$ is a non-unique minimizer. Then, there exists another minimizer $\theta' \in \R$ such that $\theta' \neq \theta^*$. Let $\alpha \in (0,1)$, and define $\theta_\alpha = \alpha \theta^* + (1-\alpha)\theta' \in \R$. But then,
    \begin{align*}
        \sup_{P \in \mathcal P} \mathbb E_{Z \sim P} [ \ell (\theta_\alpha, Z) ] &< \sup_{P \in \mathcal P} \mathbb E_{Z \sim P} [  \alpha \ell (\theta^*, Z) + (1-\alpha) \ell(\theta',Z)]\\
        &= \sup_{P \in \mathcal P}  \alpha \mathbb E_{Z \sim P} [  \ell (\theta^*, Z)] +  (1-\alpha) \mathbb E_{Z \sim P} [  \ell(\theta',Z)]\\
        &\le \sup_{P \in \mathcal P}  \alpha \mathbb E_{Z \sim P} [  \ell (\theta^*, Z)] +  \sup_{P \in \mathcal P} (1-\alpha) \mathbb E_{Z \sim P} [  \ell(\theta',Z)] \\
        &= \min_{\theta \in \R} \sup_{P \in \mathcal P} \mathbb E_{Z \sim P} [  \ell(\theta',Z)]\,.\qedhere
    \end{align*}
\end{proof}

\begin{remark}[$\Gamma$-minimax and DRO]\label{remarkGammaMinimax}
    A special case of an elicitable IP-property $f$ is given by the loss function $\ell$ of a decision maker, with $\R$ a finite set of possible decisions. In such a setting, $f$ is the corresponding $\Gamma$-minimax---or equivalently DRO---decision rule (\cf \cite{troffaes2007decision, rahimianFrameworksResultsDistributionally2022}). Given an imprecise probability $\P$, \eg a credence about nature's state, $f(\P)$ is the decision which minimizes the worst-case risk. 
	
	When $\ell$ is a strictly proper scoring rule for forecasts, $f(\mathcal P)$ can be interpreted as the set of forecasts which are $\Gamma$-minimax admissible under $\mathcal P$ (as first introduced in \cite[Definition~2.1]{schervish2025elicitation}).
\end{remark}
Naturally, IP-properties strictly extend standard (precise) properties. Every standard property can be written as an IP-property using singleton IP sets. In the other direction, every IP-property implies a certain standard property, called the \emph{precise restriction} $\hat{f} : \mathbb P \to 2^\R$, which is defined as
\begin{align*}
    \hat{f}(P) &= f(\{ P\}),
\end{align*}
where $\mathbb P = \{ P \in \Delta (\Z) \mid \{ P \} \in \scP \}$.

\subsection{Some necessary conditions for elicitability}

Equipped with these definitions we can approach the central question posed in the introduction: What can a model ``minimax-learn'' in a multi-distribution setup? To this end, we provide a first set of necessary conditions that an IP-property has to satisfy in order to be elicitable---i.e. in order to be ``minimax-learnable''. We begin by restricting our attention to so-called \emph{full} IP-properties. However, we will show later (see Remark~\ref{remarkExtend}) that this restriction is without loss of generality for elicitable properties.

\begin{definition}[Full IP-property]
\label{def:full IP-property}
    We say that an IP-property $f : \scP \to 2^\R$ is \emph{full} if all of the following conditions are satisfied:
    \begin{enumerate}
        \item If $\P \in \scP$, then $\co \P \in \scP$, where $\co \P$ denotes the closed,\footnote{For the purpose of defining the closure, we assume that there exists a dual pair $\langle X, X'\rangle$ \citep[Definition 5.90]{aliprantis2006infinite} of two vector spaces $X$ and $X'$ such that $\Delta(\Z) \subseteq X'$ with bilinear mapping $\langle \cdot, \cdot \rangle \colon X \times X' \rightarrow \mathbb{R}$. The closure is taken with respect to the weakest topology which makes all linear mappings $x \mapsto \langle x, x'\rangle \in \mathbb{R}$ for all $x' \in X'$ continuous. In case that $\Z$ is finite, the closure is simply taken with respect to the standard Euclidean topology induced on the simplex $\Delta(\Z)$.} convex hull of $\P$.
        \item The convex combination of IPs in a level set of $f$ forms an element in the domain $\scP$. That is, for all $\P, \Q \in \LIP_\theta$ and $\alpha \in [0,1]$, the IP 
    \[\alpha \P + (1-\alpha) \Q = \{ \alpha P + (1-\alpha) Q : P \in \P, Q \in \Q\}\,,\]
        is in $\scP$.
        \item The union of two IPs in a level set of $f$ also forms an element in the domain $\scP$. That is, if $\P, \Q \in \LIP_\theta$, then $\P \cup \Q \in \scP$.
    \end{enumerate}
\end{definition}

\begin{prop}[Necessary conditions for elicitability]
\label{prop:Necessary Conditions for Elicitability}
    Let $f\colon \scP \to 2^\R$ be a full, elicitable IP-property. Then, the following four statements hold:
    \begin{enumerate}[label=(\Roman*)]
        \item \label{necessaryCondition1} For all $\P \in \scP$, $f(\P) = f(\co \P)$.
        \item \label{necessaryCondition2} The level sets of $f$ are convex; \ie for all $\P, \Q \in \LIP_\theta$ and $\alpha \in [0,1]$, the IP 
    \[\alpha \P + (1-\alpha) \Q = \{ \alpha P + (1-\alpha) Q : P \in \P, Q \in \Q\}\]
    is also in $\LIP_\theta$.
        \item \label{necessaryCondition3} The level sets of $f$ are closed under arbitrary unions. That is, 
        if $\{\P_i\}_{i \in I} \subset \LIP_\theta$, then $\bigcup_{i \in I} \P_i \in \LIP_\theta$.
        \item \label{necessaryCondition4} The precise restriction $\hat{f}$ of $f$ is elicitable.
    \end{enumerate}
\end{prop}
\begin{proof}
    \begin{enumerate}[label=(\Roman*)]
        \item By linearity and continuity of $P \mapsto \mathbb{E}_{Z \sim P}[\ell(\theta, Z)]$ we can write (\cf \citep[Theorem 7.51]{aliprantis2006infinite}),
        \begin{align*}
            \argmin_{\theta \in \R} \sup_{P \in \P} \mathbb{E}_{Z \sim P}[\ell(\theta, Z)] = \argmin_{\theta \in \R} \sup_{P \in \co \P} \mathbb{E}_{Z \sim P}[\ell(\theta, Z)].
        \end{align*}
        Hence, $\theta^* \in f(\P)$ if and only if $\theta^* \in f(\co \P)$.
        \item Let $\P, \Q \in \scP$ be such that 
        $\theta^* \in f(\P)$ and $\theta^* \in f(\Q)$. Fixing some $\alpha \in [0,1]$, for all $\theta \in \R$, we have 
    \begin{align*}
    \sup_{D \in \alpha \Q + (1-\alpha)\P}\mathbb{E}_{Z \sim D}[\ell(\theta^*, Z)] &= \alpha \sup_{Q \in \Q} \mathbb{E}_{Z \sim Q}[\ell(\theta^*, Z)] +(1- \alpha) \sup_{P \in \P}\mathbb{E}_{Z \sim P}[\ell(\theta^*, Z)]\\
    &\le \alpha \sup_{Q \in \Q} \mathbb{E}_{Z \sim Q}[\ell(\theta, Z)] +(1- \alpha) \sup_{P \in \P}\mathbb{E}_{Z \sim P}[\ell(\theta, Z)]\\
    &= \sup_{D \in \alpha \Q + (1-\alpha)\P}\mathbb{E}_{Z \sim D}[\ell(\theta, Z)].
    \end{align*}

    \item 
	Let $\{\P_{i}\}_{i \in I} \subset \scP$ be such that 
	$\theta^* \in f(\P_i)$ for all $i \in I$, and write $\P_{\cup}$ for $\bigcup_{i \in I} \P_i$. Then Condition~\ref{necessaryCondition3} follows because, for all $\theta \in \R$,
	\begin{align*}
		\sup_{P \in \P_\cup} \mathbb E_{Z \sim P} [ \ell (\theta^*, Z) ] &= \sup_{i \in I} \sup_{P \in \P_i} \mathbb E_{Z \sim P} [ \ell (\theta^*, Z) ] \\
		&\le \sup_{i \in I}  \sup_{P \in \P_i} \mathbb E_{Z \sim P} [ \ell (\theta, Z) ] \\
		&= \sup_{P \in \P_\cup} \mathbb E_{Z \sim P} [ \ell (\theta, Z) ].
	\end{align*}

    \item This statement follows immediately from the elicitability of $f$.\qedhere 
    \end{enumerate}
\end{proof}
Note that the last condition means that, given a full elicitable IP property $f$, the necessary and sufficient conditions for standard elicitable properties have to hold for the restriction $\hat{f}$, \ie \cite[Theorem 1]{lambert2019elicitation} and \cite[Theorem 5]{steinwart2014elicitation}.
\begin{remark}\label{remarkExtend}
    If an IP-property is elicitable, then one can, without loss of generality, extend the domain $\scP$ by taking the closed convex hull of elements in $\scP$ (Proposition~\ref{prop:Necessary Conditions for Elicitability}, Condition~\ref{necessaryCondition1}), and closure with respect to convex combinations (Proposition~\ref{prop:Necessary Conditions for Elicitability}, Condition~\ref{necessaryCondition2}) and set unions (Proposition~\ref{prop:Necessary Conditions for Elicitability}, Condition~\ref{necessaryCondition3}) of elements in each level set.
    Hence, an elicitable IP-property can always be extended to a full elicitable property.
\end{remark}

\begin{remark}
    The level sets of an elicitable IP-property $f$ are also closed under taking the full convex combination of elements. That is, if $\P, \Q \in \LIP_\theta$ then 
        \[[\P, \Q] = \{ \alpha P + (1-\alpha) Q : P \in \P, Q \in \Q, \alpha \in [0,1]\},\]
        is also in $\LIP_\theta$. This follows by Conditions~\ref{necessaryCondition1} and~\ref{necessaryCondition3} of Proposition~\ref{prop:Necessary Conditions for Elicitability}, and the observation that $\co (\P \cup \Q) = \co [\P, \Q]$, because $\P \cup \Q \subseteq [\P, \Q] \subseteq \overline{\operatorname{co}} \P \cup \Q$.
\end{remark}

\begin{example}
\label{ex:level sets are not closed under intersection}
    As the level sets of an elicitable IP-property are closed under arbitrary union, a natural question is whether the level sets are also closed under intersection. Let us provide a counterexample. As this example requires the computation of the IP-properties the reader might first go through Example~\ref{ex:squared loss elicits the mean of maximum variance distribution} to get a better understanding why the argument holds.
    Consider the outcome set $\Z = \{ 0,1,2\}$. We denote probabilities $p \in \Delta(\Z)$ as $(p_0, p_1, p_2) \in [0,1]^3$. Let $f(\P)$ be the mean of the maximum variance distribution in $\P$. As we show in Example~\ref{example:list of elicited bayes pairs}, $f$ is elicited by the loss $\ell(\theta, z) = (\theta - z)^2$. 
    Then, for $\P = \{ (1,0,0), (0.5, 0, 0.5)\}$,
    \begin{align*}
        f(\P) &= 1,
    \end{align*}
    the mean of $p = (0.5, 0, 0.5)$ which is the maximum variance distribution in $\co \P$ (\cf Proposition~\ref{prop:necessary condition of IP-elicitability via Bayes pair}). For $\Q = \{ (1,0,0), (0.25, 0.5, 0.25)\}$, we obtain,
    \begin{align*}
        f(\Q) &= 1,
    \end{align*}
    the mean of $p = (0.25, 0.5, 0.25)$ which is the maximum variance distribution in $\co \Q$ (\cf Proposition~\ref{prop:necessary condition of IP-elicitability via Bayes pair}). However, it is easy to see that,
    \begin{align*}
        f(\P \cap \Q) = f( \{(1,0,0)\} ) = 0.
    \end{align*}
    This example is illustrated in Figures~\ref{fig:mean on simplex} and~\ref{fig:variance on simplex}.
    \begin{figure}
    \centering
    \begin{subfigure}{0.4\textwidth}
        \includegraphics[width=\textwidth]{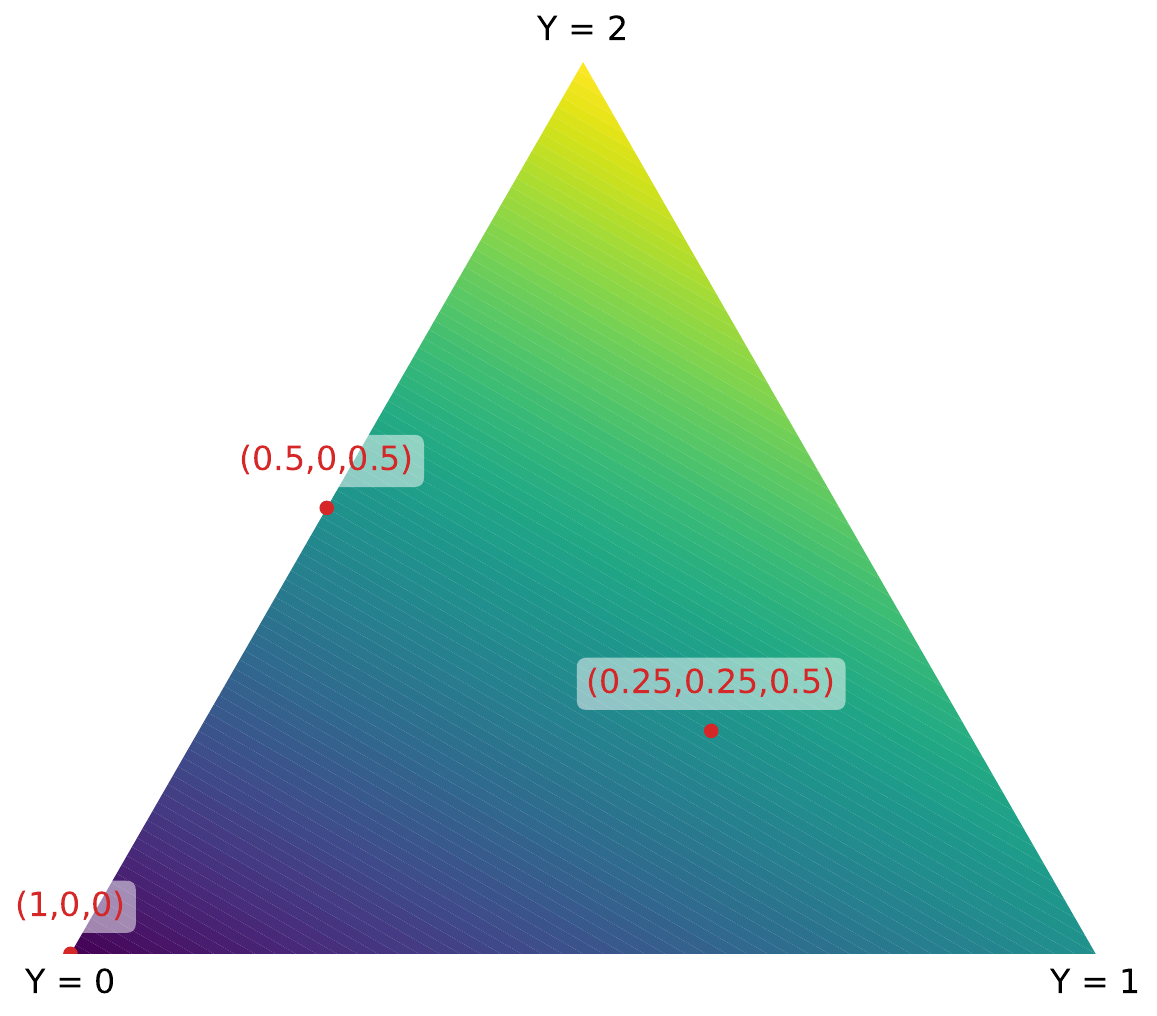}
        \caption{Mean on simplex. Distributions marked in red are taken from Example~\ref{ex:level sets are not closed under intersection}.}
        \label{fig:mean on simplex}
    \end{subfigure}
    \hfill
    \begin{subfigure}{0.4\textwidth}
        \includegraphics[width=\textwidth]{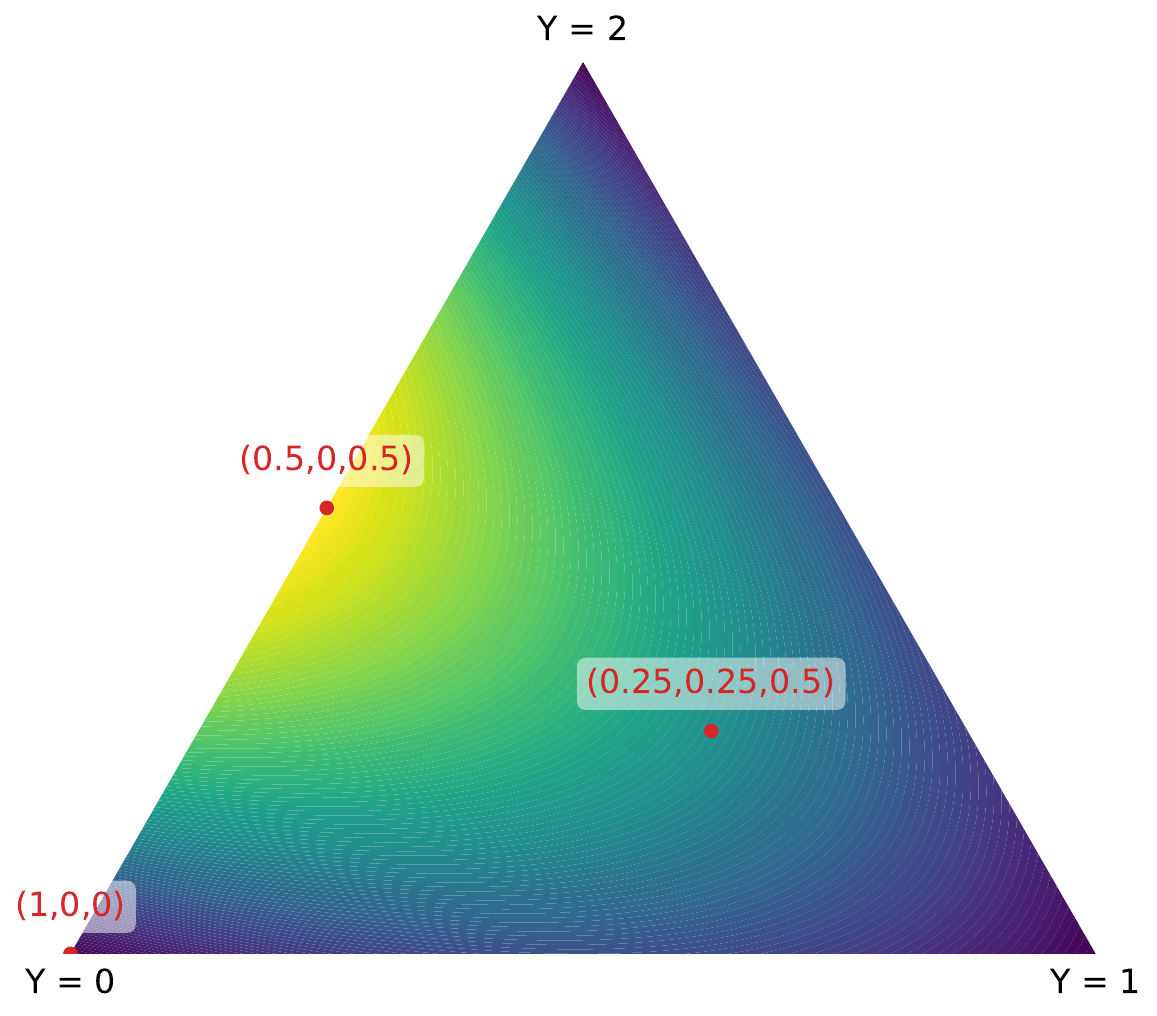}
        \caption{Variance on simplex. Distributions marked in red are taken from Example~\ref{ex:level sets are not closed under intersection}.}
        \label{fig:variance on simplex}
    \end{subfigure}
    \end{figure}
\end{example}
Proposition~\ref{prop:Necessary Conditions for Elicitability} provides a list of necessary conditions for a property to be elicitable. In other words, only if a property fulfills these conditions it can be learned in a multi-distribution setup. 
Towards a characterization of IP-elicitability we take a more detailed look on the minimax-optimization within the definition of IP-elicitation (Definition~\ref{def:elicitable IP-property}).

\subsection{Other necessary conditions via minimax theorems}
\label{sec:Necessary conditions via minimax-theorems}
Elicitable IP-properties' minimax objective implies additional conditions that such properties have to fulfill. We show that the IP-property $f$ elicited by a loss is related to the Bayes pair $(\Theta, L_\Theta)$ corresponding to that loss. More specifically, $f(\P)$ is ``equal'' (ignoring for now complications due to $f$ and $\Theta$ being set-valued) to $\Theta(P^*)$, where $P^* = \argmax_{P \in \P} L_\Theta(P)$. In other words, $f(\P)$ is the minimizer of the risk induced by the maximal Bayes risk distribution in $\P$. This reduces a robust (\ie imprecise) distributional optimization problem to a precise one.

This idea is not entirely new and can be made formal under several different technical restrictions (see \citep{grunwald2004game, frohlich2024scoring, schervish2025elicitation} or \citep[page 16]{rahimianFrameworksResultsDistributionally2022}).
To ease the technical burden, we assume that $\Z$ is finite for the remainder of the paper and that $\Delta(\Z)$ is equipped with the Euclidean topology.
\begin{prop}
    \label{prop:necessary condition of IP-elicitability via Bayes pair}
    Let $\Z$ be finite.
    Fix an imprecise probability $\P \in \scP$ which is closed and convex.
    If the IP-property $f : \scP \to 2^\R$ is bounded elicitable---\ie, elicited by a bounded loss function $\ell : \R \times \Z \to [0, B]$---then 
    \[f(\P) \subseteq \bigcap_{P^* \in \P^*} \Theta(P^*)\,,\]
    where $(\Theta, L_\Theta)$ is the Bayes pair on $\mathbb{P} := \bigcup_{\P \in \scP} \P$ induced by $\ell$ and $\P^* = \argmax_{P \in \P} L_\Theta(P)$.
\end{prop}

\begin{proof}
    Assume without loss of generality that $f(\P) \neq \emptyset$, as otherwise the proposition's conclusion holds trivially.
    For every $\theta^* \in f(\P)$ it holds,
    \begin{align*}
        \sup_{P \in \P} \mathbb{E}_{Z \sim P}[\ell(\theta^*, Z)] \le \min_{\theta \in \R} \sup_{P \in \P} \mathbb{E}_{Z \sim P}[\ell(\theta, Z)]\,.
    \end{align*}
    Such $\theta^*$ are called \emph{robust Bayes} in \citep{grunwald2004game}.
    Furthermore, the set,
    \begin{align*}
       \P^* := \argmax_{P \in \P} L_\Theta(P)\,,
    \end{align*}
    which consists of distribution $P^*\in \P^*$, such that,
    \begin{align*}
        \inf_{\theta \in \R} \mathbb{E}_{Z \sim P^*}[\ell(\theta, Z)] \ge \sup_{P \in \P} \inf_{\theta \in \R} \mathbb{E}_{Z \sim P^*}[\ell(\theta, Z)]\,,
    \end{align*}
    is non-empty \citep[Theorem 5.1]{grunwald2004game}.
    As $P^*$ maximizes the Bayes risk, it is called a \emph{maximum entropy distributions} in \citep{grunwald2004game} referring to the generalization of Shannon's entropy as Bayes risk of the cross-entropy loss. Intuitively speaking, the maximum entropy distributions are the distributions on which it is ``most difficult'' to minimize the loss. In particular, $\theta^*$ minimizes the risk against all distributions in $\P$, including all maximum entropy distributions collected in $\P^*$. It follows, \citep[Lemma 4.1 and Theorem 4.1]{grunwald2004game}, that the robust Bayes $\theta^*$ is \emph{Bayes} against all elements in $\P^*$, \ie, for all $P^* \in \P^*$
    \begin{align*}
        \mathbb{E}_{Z \sim P^*}[\ell(\theta^*, Z)] \le \min_{\theta \in \R} \mathbb{E}_{Z \sim P^*}[\ell(\theta, Z)]\,.
    \end{align*}
    That is, $\theta^* \in \Theta(P^*)$ for all $P^* \in \P^*$. The statement follows.
\end{proof}
It is possible to extend the statement to infinite spaces $\Z$. For instance, Theorem 5.1 in \citep{grunwald2004game} can be replaced by Theorem 6.1 in \citep{grunwald2004game}, which then requires $\P$ to be convex, weakly closed and tight, while the loss $\ell$ needs to be bounded above and upper semicontinuous.

Note that classical general minimax theorems such as Sion's theorem \citep{sion1958general} also allow us to derive the result of Proposition~\ref{prop:necessary condition of IP-elicitability via Bayes pair} under the conditions of these theorems. However, these conditions, in particular weak compactness of the set $\P$ and lower semicontinuity of $\ell$, are often not met in DRO \citep[page 664]{kuhn2025distributionally}. For the typical setups of MDL on the other hand---where the IP-set $\P$ is usually the closed convex hull over a finite set of distributions---the conditions of classical minimax theorem are often met.

Let us guide the reader through a simple example to illustrate Proposition~\ref{prop:necessary condition of IP-elicitability via Bayes pair}.
\begin{example}
\label{ex:squared loss elicits the mean of maximum variance distribution}
    Consider the binary outcome set $\Z = \{ 0,1\}$. We denote probabilities $p \in \Delta(\Z)$ as $p \in [0,1]$. Let $f$ be a property elicited by $\ell(\theta, z) = (\theta - z)^2$. The corresponding Bayes pair is $(\Theta, L_\Theta)$. The property $\Theta$ is the mean, and $L_\Theta$ the variance (Example~\ref{Bayes pair for squared loss}). Then, for $\P = [0,0.5]$,
    \begin{align*}
        f(\P) &=\argmin_{\theta \in \mathbb R} \sup_{P \in \P} \mathbb{E}_{Z \sim P}[\ell(\theta, Z)]\\
        &=\argmin_{\theta \in \mathbb R} \sup_{p \in [0,0.5]} (1-p) \theta^2 + p (\theta - 1)^2\\
        &=\argmin_{\theta \in \mathbb R} \sup_{p \in [0,0.5]} \theta^2 - p (2\theta - 1)\\
        &=0.5.
    \end{align*}
    Observe that $f(\P) = f(\{ 0.5\})$, where $0.5$ is the maximal variance distribution in the set $\P$.
\end{example}
To support the reader's intuition we provide a short list of exemplary loss functions and the corresponding IP-properties they elicit.
\begin{example}
\label{example:list of elicited bayes pairs}
    \begin{enumerate}
        \item If $\ell(z, \theta) = (z- \theta)^2$ and $\Z = \{0, C\}$ for some constant $C > 0$, 
        then, for all $\P \in \TwoP$, $\ell$ elicits
        the mean of maximum variance distribution in $\P$ \citep[Example 1.(iii)]{embrechts2021bayes}.
        \item If $\ell(z, \theta) = |z- \theta|$ and $\Z = \{0, C\}$, 
        then, for all $\P \in \TwoP$, $\ell$ elicits
        the median of the maximum mean-median deviation distribution in $\P$ \citep[Example 1.(v)]{embrechts2021bayes}.
        \item If $\ell(z, \theta) = \theta + \frac{e^{\gamma(y - \theta)}}{\gamma}$ for some $\gamma \in (0,\infty)$ and $\Z = \{0, C\}$, 
        then, for all $\P \in \TwoP$, $\ell$ elicits
        the maximal $\gamma$-entropic risk measure of the distributions in $\P$ \citep[Example 1.(ii)]{embrechts2021bayes}.
    \end{enumerate}
\end{example}
The following corollary follows immediately from Proposition~\ref{prop:necessary condition of IP-elicitability via Bayes pair}.
\begin{corollary}
\label{corollary:level sets and intersection with IP}
    Let $f : \scP \to 2^\R$ be a bounded elicitable IP-property, with loss function $\ell \colon \R \times \mathcal{Z} \rightarrow [0,B]$ and suppose all of the conditions of Proposition~\ref{prop:necessary condition of IP-elicitability via Bayes pair} are fulfilled. Fix a $\theta \in \R$ such that $\L_\theta \neq \emptyset$ is the level set of $\theta$ for the property $\Theta$ of the Bayes pair $(\Theta, L_\theta)$ induced by $\ell$. If $\L_\theta \cap \P = \emptyset$ for a convex, closed, non-empty $\P \in \scP$, then $\theta \notin f(\P)$.
\end{corollary}
\begin{proof}
    By Proposition~\ref{prop:necessary condition of IP-elicitability via Bayes pair} $f(\P) \subseteq \bigcap_{P^* \in \P^*} \Theta(P^*)$, but $\P^* \subseteq \P$ and $\L_\theta \cap \P = \emptyset$, hence $\L_\theta \cap \P^* = \emptyset$. It follows that $\theta \notin \bigcap_{P^* \in \P^*} \Theta(P^*)$, which implies the statement.
\end{proof}
A central open question is now: are the given necessary conditions also sufficient? That is, do these conditions characterize IP-elicitability? The following section answers this in the affirmative, at least in a restricted setting. For comparison, see the sufficient and necessary conditions for standard (precise) properties given in \cite[Theorem 1]{lambert2019elicitation} and \cite[Theorem 5]{steinwart2014elicitation}.

\subsection{A sufficient condition for elicitability}
\label{sec:Sufficient condition for elicitability}
To a certain extent, the necessary condition of Proposition~\ref{prop:necessary condition of IP-elicitability via Bayes pair} is close to characterizing elicitable IP-properties. Under the additional assumption that the Bayes pair assumes unique values and the property $f$ is not merely mapping to empty sets, we can show that the existence of a Bayes pair implies the elicitability of the IP-property.
\begin{prop}
    \label{prop:sufficient condition for IP-elicitability}
    Let $\Z$ be finite.
    Let $\scP \subseteq 2^{\Delta(\Z)}$ such that every $\P \in \scP$ is closed and convex.
    The IP-property $f \colon \scP \to 2^\R$ is elicitable if there exists a Bayes pair $(\Theta, L_\Theta)$ such that
    $|\Theta(P)| = 1$ for all $P \in \bigcup_{\P \in \scP} \argmax_{P \in \P} L_\Theta(P)$, and for all $\P \in \scP$, 
    $f(\P) = \Theta(P^*)$ for unique $P^* = \argmax_{P \in \P} L_\Theta(P)$. 
\end{prop}
\begin{proof}
    Let $\ell \colon \R \times \Z \to \mathbb R$ be the loss function corresponding to the Bayes pair $(\Theta, L_\Theta)$.
    Define the IP-property $f'$ on $\scP$ via elicitation by $\ell$. We show that $f' = f$.

    Fix any $\P \in \scP$.
    Note that $P^* = \argmax_{P \in \P} L_\Theta(P)$ is unique. In the words of \citep{grunwald2004game}, $P^*$ is a unique distribution maximizing the generalized entropy $L_\Theta$.
    Furthermore, $\Theta(P^*) \in \R$ maps to a unique value.
    Hence, Corollary 4.1 in \citep{grunwald2004game} applies.
    Because $\theta^* := \Theta(P^*) \in \R$ is \emph{Bayes} against $P^*$, \ie,
     \begin{align*}
        \mathbb{E}_{Z \sim P^*}[\ell(\theta^*, Z)] \le \min_{\theta \in \R} \mathbb{E}_{Z \sim P^*}[\ell(\theta, Z)]\,.
    \end{align*}
    it follows that $\theta^*$ is \emph{robust Bayes} against $\P$, \ie,
     \begin{align*}
        \sup_{P \in \P} \mathbb{E}_{Z \sim P}[\ell(\theta^*, Z)] \le \min_{\theta \in \R} \sup_{P \in \P} \mathbb{E}_{Z \sim P}[\ell(\theta, Z)]\,.
    \end{align*}
    In other terms, $f'(\P) \supseteq \Theta(P^*)$. The reverse implication holds as well, \ie, a robust Bayes $\theta^*$ against $\P$ is Bayes against $P^*$, \ie, $f'(\P) \subseteq \Theta(P^*)$ \citep[Theorem 4.1]{grunwald2004game}.
    
    As the argument holds for any $\P \in \scP$, $f' = f$. That is, $f$ is elicited by $\ell$.
\end{proof}
The conditions in Proposition~\ref{prop:sufficient condition for IP-elicitability} can be non-trivially fulfilled, \eg, by Example~\ref{ex:squared loss elicits the mean of maximum variance distribution}.

\section{Conclusion and open questions}
In this paper, we provide a list of necessary conditions and a sufficient condition for the elicitability of an IP-property. As we argued in the introduction, only elicitable IP-properties \emph{can} be ``minimax-learned'' in multi-distribution setups. In particular, Proposition~\ref{prop:necessary condition of IP-elicitability via Bayes pair} shows the limitations of elicitable IP-properties. For instance, the ``uncertainty'' encoded in the size of the set $\P$ cannot be captured by any elicitable IP-property $f$. The IP-property $f$ only depends on a single element of the set $\P$.

Nevertheless, this work leaves a series of interesting open questions unanswered. What is a full characterization of elicitable properties? What is the analogue of identifiability---\ie, properties which can be identified through a minimizer condition \citep{steinwart2014elicitation}---in the imprecise case? What is the set of loss functions which elicit the same (IP-)property and what are the corresponding possible Bayes risks?

\vspace{1em}
\subsection*{Acknowledgements}
\addcontentsline{toc}{section}{Acknowledgements}

We thank the anonymous reviewer of our poster submission to ISIPTA 2025 for their feedback, and for drawing our attention to the link between this work and the topic of structural judgments. 
JB gratefully acknowledges partial financial support from the Australian-American Fulbright Commission and the Kinghorn Foundation. 
Thanks to the International Max Planck Research School for Intelligent Systems (IMPRS-IS) for supporting RD. RD was funded by the Deutsche Forschungsgemeinschaft (DFG, German Research Foundation) under Germany’s Excellence Strategy -- EXC number 2064/1 -- Project number 390727645; he was also supported by the German Federal Ministry of Education and Research (BMBF): T\"ubingen AI Center.

\begingroup
\sloppy
\printbibliography

@inproceedings{steinwart2014elicitation,
  title={Elicitation and identification of properties},
  author={Steinwart, Ingo and Pasin, Chlo{\'e} and Williamson, Robert and Zhang, Siyu},
  booktitle={Proceedings of the 27th Conference on Learning Theory},
  pages={482--526},
  year={2014},
  publisher={Proceedings of Machine Learning Research}
}

@incollection{mirandaStructuralJudgements2014,
  title = {Structural Judgements},
  booktitle = {Introduction to Imprecise Probabilities},
  author = {Miranda, Enrique and family=Cooman, given=Gert, prefix=de, useprefix=true},
  date = {2014-05-09},
  series = {Wiley {{Series}} in {{Probability}} and {{Statistics}}},
  pages = {56--78},
  publisher = {John Wiley \& Sons, Ltd},
}

@article{rahimianFrameworksResultsDistributionally2022,
  title = {Frameworks and results in distributionally robust optimization},
  author = {Rahimian, Hamed and Mehrotra, Sanjay},
  year = {2022},
  journal = {Open Journal of Mathematical Optimization},
  volume = {3},
  pages = {1--85}
}

@article{brierVerificationForecastsExpressed1950,
  title = {Verification of Forecasts Expressed in Terms of Probability},
  author = {Brier, Glenn W.},
  year = {1950},
  month = jan,
  journal = {Monthly Weather Review},
  volume = {78},
  number = {1},
  pages = {1--3},
}

@report{lambert2019elicitation,
  title={Elicitation and evaluation of statistical forecasts},
  author={Lambert, Nicolas S.},
  year={2019},
  REMOVEDFROMABSTRACTurl={https://ai.stanford.edu/~nlambert/papers/elicitation_statistical_forecasts.pdf},
  type={Working Paper}
}

@book{osband1985providing,
  title={Providing Incentives for Better Cost Forecasting},
  author={Osband, Kent Harold},
  year={1985},
  publisher={PhD Thesis, University of California}
}

@article{troffaes2007decision,
  title={Decision making under uncertainty using imprecise probabilities},
  author={Troffaes, Matthias C. M.},
  journal={International Journal of Approximate Reasoning},
  volume={45},
  number={1},
  pages={17--29},
  year={2007},
  publisher={Elsevier}
}

@book{schechter1997handbook,
  title={Handbook of analysis and its foundations},
  author={Schechter, Eric},
  year={1997},
  publisher={Academic Press}
}

@article{sion1958general,
  title = {On general minimax theorems},
  author = {Sion, Maurice},
  year = {1958},
  journal = {Pacific Journal of Mathematics},
  volume = {8},
  number = {1},
  pages = {171--176}
}

@book{aliprantis2006infinite,
  title={Infinite dimensional analysis: {{A}} hitchhiker's guide},
  author={Aliprantis, Charalambos D. and Border, Kim C.},
  year={2006},
  publisher={Springer Science \& Business Media}
}

@article{haghtalab2022demand,
  title={On-demand sampling: Learning optimally from multiple distributions},
  author={Haghtalab, Nika and Jordan, Michael and Zhao, Eric},
  journal={Advances in Neural Information Processing Systems},
  volume={35},
  pages={406--419},
  year={2022}
}

@book{augustin2014introduction,
  title={Introduction to imprecise probabilities},
  author={Augustin, Thomas and Coolen, Frank P.A. and de Cooman, Gert and Troffaes, Matthias C.M.},
  year={2014},
  publisher={John Wiley \& Sons}
}

@article{gneiting2011making,
  title={Making and evaluating point forecasts},
  author={Gneiting, Tilmann},
  journal={Journal of the American Statistical Association},
  volume={106},
  number={494},
  pages={746--762},
  year={2011},
  publisher={Taylor \& Francis}
}

@article{fissler2021forecast,
  title={Forecast evaluation of quantiles, prediction intervals, and other set-valued functionals},
  author={Fissler, Tobias and Frongillo, Rafael and Hlavinov{\'a}, Jana and Rudloff, Birgit},
  journal={Electronic Journal of Statistics},
  volume={15},
  pages={1034--1084},
  year={2021}
}

@article{brehmer2021scoring,
  title={Scoring interval forecasts: Equal-tailed, shortest, and modal interval},
  author={Brehmer, Jonas R. and Gneiting, Tilmann},
  journal={Bernoulli},
  volume={27},
  number={3},
  pages={1993--2010},
  year={2021}
}

@article{singh2025truthful,
  title={Truthful elicitation of imprecise forecasts},
  author={Singh, Anurag and Chau, Siu Lun and Muandet, Krikamol},
  journal={arXiv preprint arXiv:2503.16395},
  year={2025}
}

@article{frongillo2021elicitation,
  title={Elicitation complexity of statistical properties},
  author={Frongillo, Rafael M. and Kash, Ian A.},
  journal={Biometrika},
  volume={108},
  number={4},
  pages={857--879},
  year={2021},
  publisher={Oxford University Press}
}

@article{ziegel2016coherence,
  title={Coherence and elicitability},
  author={Ziegel, Johanna F.},
  journal={Mathematical Finance},
  volume={26},
  number={4},
  pages={901--918},
  year={2016},
  publisher={Wiley Online Library}
}

@inproceedings{konek2023evaluating,
  title={Evaluating imprecise forecasts},
  author={Konek, Jason},
  booktitle={International Symposium on Imprecise Probability: Theories and Applications},
  pages={270--279},
  year={2023},
  organization={PMLR}
}

@report{frohlich2024scoring,
  title={Scoring rules and calibration for imprecise probabilities},
  author={Fr{\"o}hlich, Christian and Williamson, Robert C.},
  type={arXiv preprint arXiv:2410.23001},
  year={2024}
}

@article{grunwald2004game,
  title={Game theory, maximum entropy, minimum discrepancy and robust {{Bayesian}} decision theory},
  author={Gr{\"u}nwald, Peter D. and Dawid, A. Philip},
  journal={The Annals of Statistics},
  volume={32},
  number={4},
  pages={1367--1433},
  year={2004}
}

@article{embrechts2021bayes,
  title={Bayes risk, elicitability, and the expected shortfall},
  author={Embrechts, Paul and Mao, Tiantian and Wang, Qiuqi and Wang, Ruodu},
  journal={Mathematical Finance},
  volume={31},
  number={4},
  pages={1190--1217},
  year={2021},
  publisher={Wiley Online Library}
}

@inproceedings{schervish2025elicitation,
  title={Elicitation for sets of probabilities and distributions},
  author={Schervish, Mark J. and Seidenfeld, Teddy and Gong, Ruobin and Kadane, Joseph B. and Stern, Rafael B.},
  booktitle = {Proceedings of the {{Fourteenth International Symposium}} on {{Imprecise Probability}}: {{Theories}} and {{Applications}}},
  pages={242--251},
  year={2025},
  volume={290},
  publisher = {Proceedings of Machine Learning Research}
}

@article{kuhn2025distributionally,
  title={Distributionally robust optimization},
  author={Kuhn, Daniel and Shafiee, Soroosh and Wiesemann, Wolfram},
  journal={Acta Numerica},
  volume={34},
  pages={579--804},
  year={2025},
  publisher={Cambridge University Press}
}
\addcontentsline{toc}{section}{\refname}
\endgroup

\end{document}